\documentclass[runningheads]{llncs}

\usepackage{eccv}
\usepackage{eccvabbrv}

\usepackage{graphicx}
\usepackage{booktabs}

\usepackage[accsupp]{axessibility}  
\usepackage{wrapfig}

\usepackage{hyperref}
\usepackage{xr-hyper}
\usepackage{orcidlink}

\begin{document}

\title{PhysNeXt: Next-Generation Dual-Branch Structured Attention Fusion Network for Remote Photoplethysmography Measurement} 

\titlerunning{PhysNeXt: Dual-Branch Attention Network for rPPG}

\author{
Junzhe Cao\inst{1,2}$^{\dagger}$
\and
Bo Zhao\inst{2}$^{\dagger}$
\and
Zhiyi Niu\inst{2}
\and
Dan Guo\inst{3}
\and
Yue Sun\inst{4}
\and
Haochen Liang\inst{2}
\and
Yong Xu\inst{1}$^{*}$
\and
Zitong Yu\inst{2}$^{*}$
}

\authorrunning{J. Cao et al.}

\institute{
Harbin Institute of Technology, Shenzhen \and
Great Bay University \and
Hefei University of Technology \and
Macao Polytechnic University
}

\maketitle
\begingroup
\renewcommand\thefootnote{}\footnotetext{$^{*}$ Corresponding authors.}
\footnotetext{$^{\dagger}$ These authors contributed equally to this work.}
\endgroup

\begin{abstract}
  Remote photoplethysmography (rPPG) enables contactless measurement of heart rate and other vital signs by analyzing subtle color variations in facial skin induced by cardiac pulsation. Current rPPG methods are mainly based on either end-to-end modeling from raw videos or intermediate spatial–temporal map (STMap) representations. The former preserves complete spatiotemporal information and can capture subtle heartbeat-related signals, but it also introduces substantial noise from motion artifacts and illumination variations. The latter stacks the temporal color changes of multiple facial regions of interest into compact two-dimensional representations, significantly reducing data volume and computational complexity, although some high-frequency details may be lost. To effectively integrate the mutual strengths, we propose PhysNeXt, a dual-input deep learning framework that jointly exploits video frames and STMap representations. By incorporating a spatio-temporal difference modeling unit, a cross-modal interaction module, and a structured attention–based decoder, PhysNeXt collaboratively enhances the robustness of pulse signal extraction. Experimental results demonstrate that PhysNeXt achieves more stable and fine-grained rPPG signal recovery under challenging conditions, validating the effectiveness of joint modeling of video and STMap representations. The codes will be released.
  \keywords{Remote Heart Rate Measurement \and rPPG \and Dual-Branch}
\end{abstract}

\section{Introduction}
\label{sec:intro}
Heart rate is a key physiological indicator for assessing cardiovascular health, mental stress, and overall well-being, and it is widely used in applications such as telemedicine, driving safety monitoring, and human–computer interaction\cite{yang2025selfnicu,mcduff2014remotecognitive_stress,choi2025mmdrive,wang2025physdrive}. Traditional contact-based measurements, such as electrocardiography (ECG) and fingertip pulse oximeters, are highly accurate. However, they require electrodes or sensors to be attached to the subject, which restricts freedom of movement and reduces wearing comfort. In contrast, remote photoplethysmography (rPPG) technology captures the subtle color changes of facial skin with the heartbeat cycle using a regular RGB camera, and can measure heart rate and other vital signs without direct contact with the skin. rPPG has the advantages of being non-invasive, low-cost and easy to integrate. In recent years, with the development of computer vision and deep learning technologies, a large number of studies have continuously promoted the improvement of rPPG methods\cite{verkruysse2008remoteorigin,poh2010non,poh2010advancements,de2013robust,li2014remote,tulyakov2016self,yu2019remote,niu2020video-cvd}.

The basic principle of rPPG is to analyze the color changes of the facial skin region in consecutive video frames and extract the implicit heart rhythm signal\cite{chen2018deepphys}. According to the input form, the deep learning-based rPPG method can be mainly divided into two categories: one is an end-to-end method that uses the original facial video sequence as the network input\cite{yu2019remotephysnet,yu2022physformer,sun2024Contrast-phys+,MTTSliu2020multi,zou2025rhythmformer}, and the other is to preprocess the video into a spatial-temporal map (STMap) before inputting it into the network\cite{niu2019rhythmnet,liu2024rppg-mae,niu2020video-cvd,lu2021dual-gan}.
\begin{figure}[t]
  \centering
  \includegraphics[width=\linewidth]{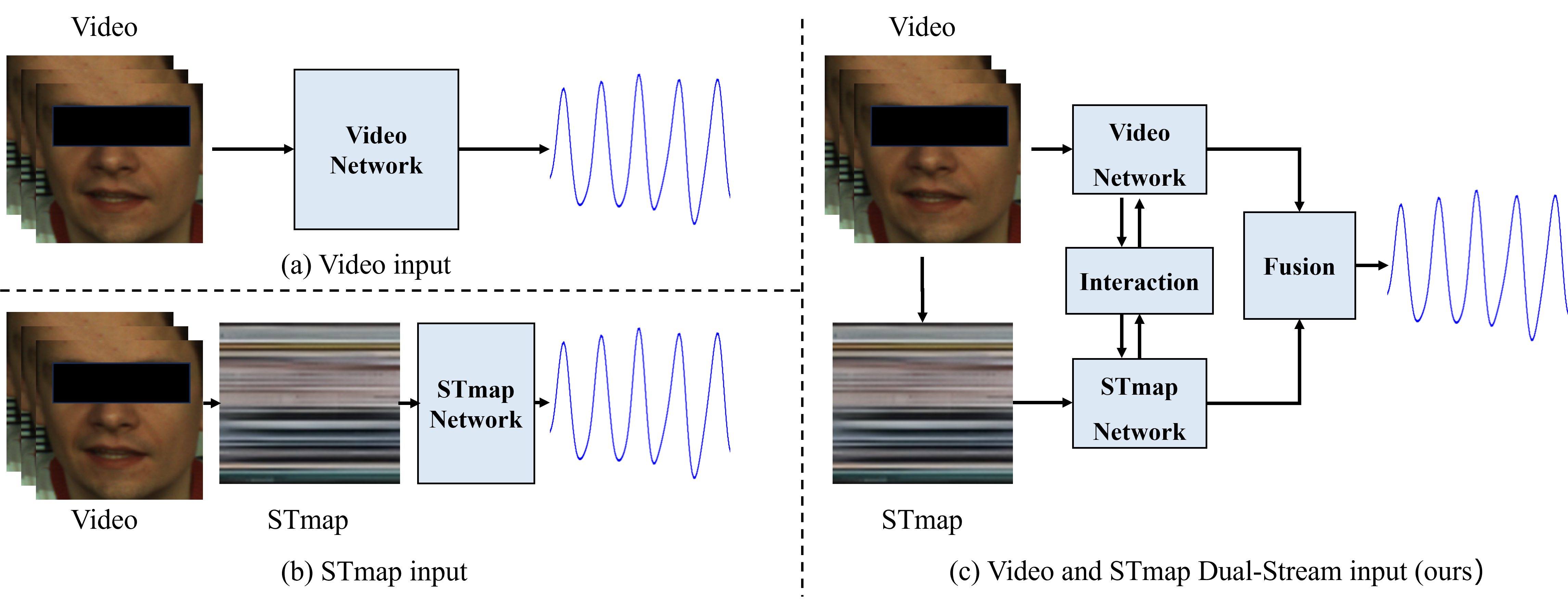}
  \caption{The comparison between (a) Video input, (b) STMap input, and (c) our proposed Video and STMap Dual-Stream input network.}
  \label{fig:fig1}
  \vspace{-2em}
\end{figure}
The video retains complete spatiotemporal information and can capture subtle color changes caused by heartbeats, but it also brings a lot of noise such as motion artifacts and lighting changes. In contrast, the STMap method first selects multiple regions of interest (ROIs) for faces and stacks the color changes of these regions along the time axis into a two-dimensional image, providing a more compact structured representation for the model. This preprocessing greatly reduces the data volume compared with raw video and enables 2D convolutional networks to learn the temporal characteristics of pulse signals more efficiently, thereby lowering the computational complexity. However, the compression process of the STMap method also loses some high-frequency details in the original video, which may lead to a decrease in the accuracy of pulse waveform recovery. Therefore, how to effectively integrate the advantages of these two types of information structures in a single model remains a question worthy of research.

To address this challenge, we propose a dual-stream deep learning framework based on video and STMap inputs, which jointly processes raw video frames and corresponding STMap representations to enhance the accuracy and robustness of pulse signal extraction. In the STMap branch, a Spatio-Temporal Difference Modeling Unit is introduced to capture subtle heartbeat-related skin color variations through temporal and spatial differences. 
Furthermore, we design a bidirectional interaction mechanism between the video and STMap encoders based on frequency-domain waveform matching and confidence-gated modulation. This module aligns features from the two modalities using frequency-domain correlation and uses confidence gating to control the direction and strength of information flow, enabling reliable cross-modal feature complementarity.
In the decoding stage, we design a structured attention mechanism that introduces a global state token and multiple learnable query tokens. Masked multi-head attention is used to control the interactions among different tokens, enabling fine-grained fusion of multi-source information and improving the accuracy of rPPG waveform recovery.
Our main contributions are summarized as follows:
\begin{itemize}
\item We design a Spatio-Temporal Difference Modeling Unit in the STMap branch to explicitly enhance the modeling of heartbeat-induced skin color variations.
    
\item We propose a cross-modal interaction module that achieves reliable feature complementarity through frequency-domain waveform matching and confidence-gated modulation.
    
 \item We introduce a structured attention mechanism in the fusion decoder to further integrate multi-source signals for reliable rPPG measurement.

\item Our final model, PhysNeXt, achieves superior performance on four benchmark datasets, including both intra-dataset and cross-dataset evaluations.

\end{itemize}
\section{Related Work}
\textbf{Methods with Video Input.}
Directly using raw video as input to a neural network for physiological signal measurement is a common deep learning–based approach.
These methods~\cite{yu2019remotephysnet,sun2022contrast,chen2018deepphys,yu2022physformer,yu2023physformer++,yu2019remote} usually do not rely on explicit feature engineering or intermediate representations and have end-to-end modeling capabilities. Existing methods include various network structures, such as convolutional neural networks (CNN)\cite{chen2018deepphys,yu2019remotephysnet}, Transformer\cite{yu2022physformer,zou2025rhythmformer}, and self-supervised learning\cite{sun2022contrast,sun2024Contrast-phys+}. 
Multi-task temporal shift attention network (MTTS-CAN)~\cite{MTTSliu2020multi} introduces a temporal shift module (TSM) together with convolutional attention to enable efficient spatiotemporal modeling, allowing a single network to jointly estimate heart rate and respiratory rate in real time on mobile devices.
PhysFormer~\cite{yu2022physformer} introduces a temporal difference Transformer module, which effectively combines local details and global temporal features, significantly improving the ability to model heart rate cycle signals. 
Contrast-Phys+~\cite{sun2024Contrast-phys+} uses a 3D-CNN to produce spatiotemporal rPPG signals and applies contrastive learning on their power spectral densities.
In this framework, rPPG and GT samples from the same video are aligned, while those from different videos are separated, enabling robust unsupervised and weakly-supervised training.
However, rPPG methods that use raw video as input face several challenges: motion artifacts and illumination changes can overwhelm weak pulse signals, and the large amount of redundant spatio-temporal information in raw videos makes effective feature extraction more difficult. To address this challenge, we propose PhysNeXt, a dual-stream model that extracts complementary features from video and STMap branches and performs confidence-gated cross-modal fusion to achieve robust rPPG signal recovery.

\noindent\textbf{Methods with Spatial–Temporal Map Input.}
Another line of research employs the Spatial–Temporal Map (STMap) as an intermediate representation, transforming raw facial videos into two-dimensional image formats for physiological signal modeling~\cite{niu2019rhythmnet,niu2020video-cvd,wang2025physmle,lu2021dual-gan,liu2024rppg-mae}. STMap typically stacks the color changes of multiple regions of interest (ROIs) across time, encoding periodic physiological fluctuations as image textures. This representation is more suitable for 2D convolution and simplifies network design.
STMap was first introduced in RhythmNet~\cite{niu2019rhythmnet}, where multi-ROI facial color time series are stacked into a spatial–temporal representation and processed by a CNN–GRU network for heart rate regression.
Niu \etal ~\cite{niu2020video-cvd} convert paired face videos into multi-scale STMaps (MSTMaps) and employ a dual-encoder cross-verified autoencoder to disentangle physiological signals from non-physiological factors.
Dual-GAN~\cite{lu2021dual-gan} leverages STMaps within a coupled BVP-GAN and Noise-GAN framework to adversarially disentangle physiological signals from noise.
 rPPG-MAE~\cite{liu2024rppg-mae} proposes a masked-autoencoder-based self-supervised framework on STMaps, where masked patches are reconstructed with pixel loss and an rPPG-aware Pearson constraint, enabling the ViT encoder to learn periodic physiological priors and improve downstream HR estimation.
 However, STMap construction may lose some spatial and temporal details, limiting precise physiological signal extraction. To address this, the proposed PhysNeXt introduces a spatio-temporal difference modeling unit in the STMap branch and a structured attention mechanism in the fusion decoder for accurate and stable rPPG measurement.

\noindent\textbf{Methods with Multi-Sensor Modalities.}
Recent multimodal rPPG methods~\cite{kurihara2021nonRGB/NIR,wu2025cardiacmamba,ying2025fusionphys,ge2025evidential} combine signals from multiple sensors, such as RGB, NIR, thermal imaging, and radar, to improve robustness under challenging conditions. Kurihara \etal~\cite{kurihara2021nonRGB/NIR} proposed an adaptive RGB–NIR fusion strategy based on illumination estimation. CardiacMamba~\cite{wu2025cardiacmamba} introduced an RGB–RF fusion framework with temporal modeling and cross-modal interaction. The SATM module~\cite{liang2025spatiaSATMl} addressed spatial and temporal misalignment between video and radar. FusionPhys~\cite{ying2025fusionphys} further modeled heterogeneous modalities as unified physiological signals for adaptive fusion. 
Overall, multimodal fusion improves robustness but introduces challenges in cross-modal alignment and consistency.
\vspace{-1em}

\section{Proposed Method}
\begin{figure}[t]
  \centering
  \includegraphics[width=\linewidth]{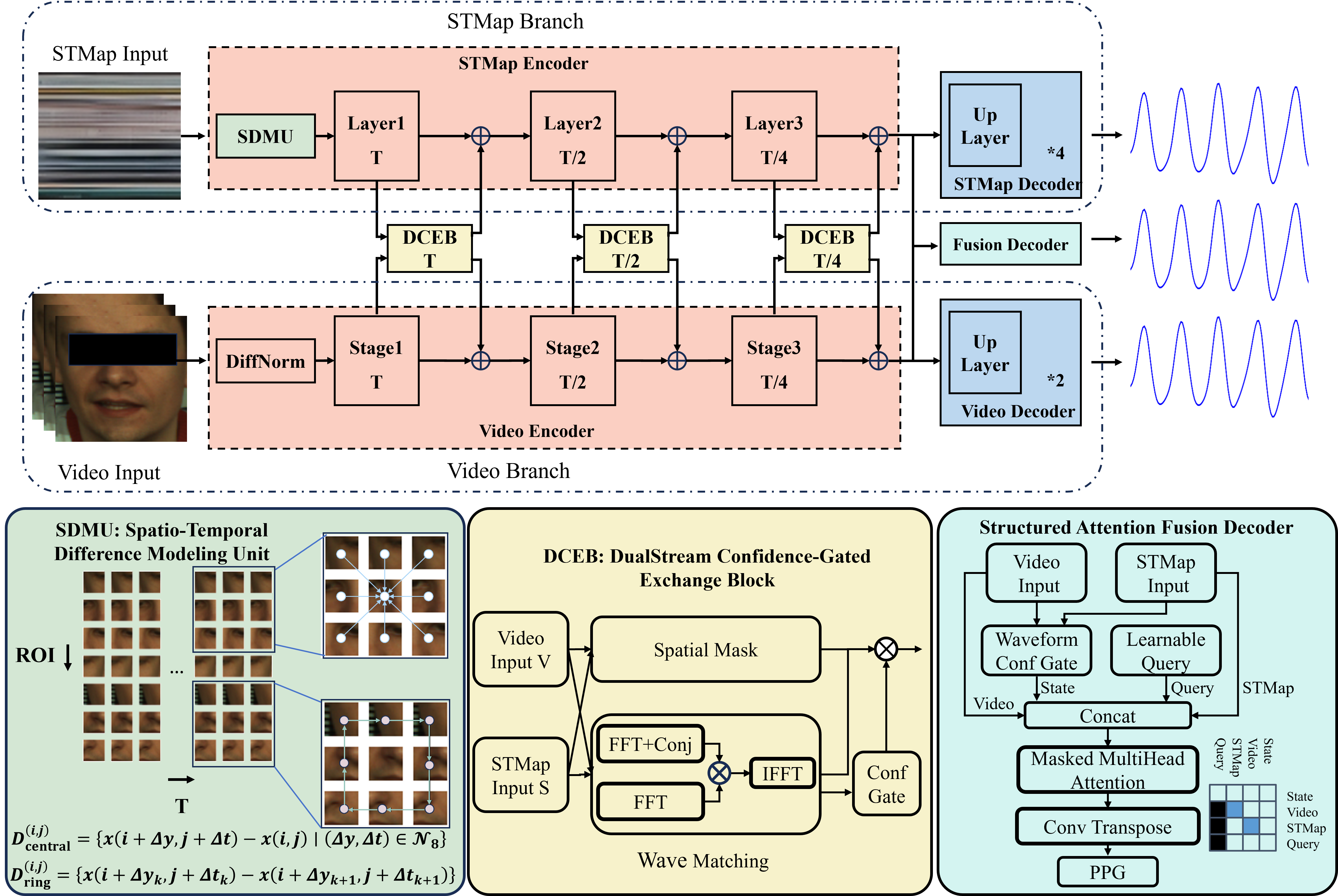}
  \caption{\textbf{The overall framework of PhysNeXt.} PhysNeXt is a dual-stream network that processes video and STMap inputs in parallel. The video and STMap branches use PhysNet and RhythmNet as their backbones, respectively. In the STMap branch, Spatio-Temporal Difference Modeling Units (SDMUs) are embedded at multiple layers to enhance spatio-temporal feature extraction. The two branches interact at multiple temporal scales through Confidence-Gated Exchange Blocks, enabling feature alignment and information exchange. Finally, a Structured Attention Fusion Decoder fuses features from both streams to predict accurate rPPG signals.}
  \label{fig:physnext}
  \vspace{-2em}
\end{figure}

As illustrated in \cref{fig:physnext}, PhysNeXt is a dual-stream network designed for remote photoplethysmography measurement. The network contains two parallel branches: one branch takes STMap~\cite{niu2019rhythmnet} as input, and the other branch takes the original video frame sequence as input. 
The input to PhysNeXt consists of a video frame sequence $V \in \mathbb{R}^{B\times 3\times T\times H\times W}$ and the corresponding STMap tensor $S \in \mathbb{R}^{B\times 3\times H'\times T}$, where $H'$ is the spatial ROI dimension and $T$ is the number of frames.
The network output is the estimated pulse signal waveform $\hat{p}(t)$,with temporal length $T$. These two branches perform feature alignment and information exchange at multiple time scales through confidence-gated interaction modules, aligning their features in both the temporal and spatial domains. Finally, a structured attention decoder fuses the two information streams, focusing on the most reliable component in each branch, and outputs a clean rPPG signal. 
The full algorithmic pipeline is detailed in  Algorithm 1 (see Appendix B).

\subsection{Spatio-Temporal Difference Modeling Unit}
\label{sec:3.2Spatio-Temporal Difference Modeling Unit}

We introduce the Spatio-Temporal Difference Modeling Unit (SDMU) to enhance the subtle variations caused by the pulse in STMap. The core operation of this module is Pixel Difference Convolution (PDC)~\cite{su2025rapid}, which aims to capture local intensity change patterns in different spatiotemporal directions.

\noindent\textbf{Central-PDC:} Calculating the difference between the current pixel and its $3\times 3$ neighborhood in 8 directions, used to measure the intensity of pixel variations along different spatial and temporal directions. Defined as follows:
\begin{equation}
D_{\text{central}}^{(i,j)}
=
\left\{
x(i+\Delta y,\, j+\Delta t) - x(i,j)
\;\middle|\;
(\Delta y,\Delta t)\in\mathcal{N}_8
\right\},
\label{eq:central_pdc}
\end{equation}
where $\mathcal{N}_8$ is a set containing 8 neighborhood offsets, covering neighbors in the space and time axes.

\noindent\textbf{Ring-PDC:} Modeling the intensity difference between adjacent pixel pairs in a clockwise order along the circumference of a $3\times 3$ neighborhood:
\begin{equation}
D_{\text{ring}}^{(i,j)}
=
\left\{
x(i+\Delta y_k,\, j+\Delta t_k) - x(i+\Delta y_{k+1},\, j+\Delta t_{k+1})
\;\middle|\;
k=1,\dots,8
\right\}.
\label{eq:ring_pdc}
\end{equation}

The two sets of difference features are weighted and combined using learnable linear transformations. We process the central difference and the ring difference using two independent convolutional weight matrices ($W_c$) and ($W_r$), then concatenate their outputs and fuse them using a $1\times 1$ convolution ($W_f$).

The entire PDC module can be formally expressed as:
\begin{equation}
y_{\text{PDC}}(i,j)
=
\sigma\!\left(
W_f \Bigl[
W_c\, D_{\text{central}}^{(i,j)}
\;;\;
W_r\, D_{\text{ring}}^{(i,j)}
\Bigr]
\right),
\label{eq:pdc_output}
\end{equation}
where $\bigl[\cdot\,;\,\cdot\bigr]$ represents the concatenation operation, and $\sigma(\cdot)$ is the non-linear activation function.

Through this differential modeling approach, PDC effectively suppresses constant background and static illumination components while amplifying subtle color changes driven by blood flow. As a result, the network is encouraged to focus on physiologically meaningful dynamic features. We embedded this module in multiple network layers of the STMap branch, including the input layer and intermediate feature layers, ensuring that the model has the ability to model time-varying changes from the initial stage.
\vspace{-1em}
\subsection{DualStream Confidence-Gated Exchange Block}
\label{sec:3.3DualStream Confidence-Gated Exchange Block}

To enable bidirectional feature exchange between the video and STMap streams, we propose a Dual-Stream Confidence-Gated Exchange Block (DCEB). The exchanged features are adaptively weighted based on their confidence.
The input to this block is a pair of feature tensors: Video branch features are denoted as $v$, and STMap branch features as $s $. 

\noindent\textbf{Spatial Alignment.} 
The exchange block generates a spatial attention map from one branch to modulate the features of the other branch.
For the STMap feature $s$, we compute its temporal average and transform it with an MLP, then add a learnable spatial template $M_0$. The result is upsampled to the video resolution and activated by a Sigmoid function to produce the attention map.

\begin{equation}
A_{b,t,h,w}^{(S \to V)}
=\sigma\!\left(\operatorname{Up}\!\left(
M_{0} + \operatorname{MLP}\!\left(\operatorname{Mean}_{t}(s)\right)
\right)\right).
\label{eq:spa_mask_s2v}
\end{equation}

This map reflects the importance of each spatial location in the video feature space with respect to the STMap representation.
Adaptive average pooling is performed over the spatial dimensions to compress it to the spatial dimension $H_s$ used by the STMap, resulting in the attention mask.
\begin{equation}
A_{b,t,h'}^{(V \rightarrow S)}
= \sigma\!\left(
\operatorname{Pool}_{H_v \times W_v \rightarrow H_s}
\!\left(
\operatorname{Conv}_{1\times1\times1}(v)
\right)
\right).
\end{equation}

Through this cross-modal attention mechanism, the two branches provide spatial guidance to each other and compensate for missing contextual information, thereby improving feature alignment and structural consistency.

\noindent\textbf{Waveform Matching.}
Waveform matching measures the temporal similarity between the video and STMap features. We first apply spatial global average pooling to each branch to obtain the temporal feature sequence:
\begin{equation}
v_{\text{global}}(t)=\frac{1}{H_vW_v}\sum_{i,j}v(t,i,j),\quad
s_{\text{global}}(t)=\frac{1}{H_s}\sum_i s(i,t)
\vspace{-0.5em}
\end{equation}

Next, we calculate the cross-correlation between these two global features in the frequency domain. Specifically, we compute the cross-correlation between $v_{\text{global}}$ and $\tilde{s}$, and between $s_{\text{global}}$ and $\tilde{v}$, as

\begin{equation}
r_{s \to v}(t)=
\mathcal{F}^{-1}\!\left(
\overline{\mathcal{F}\{v_{\text{global}}\}}
\cdot
\mathcal{F}\{\tilde{s}\}
\right)(t),\quad
r_{v \to s}(t)=
\mathcal{F}^{-1}\!\left(
\overline{\mathcal{F}\{s_{\text{global}}\}}
\cdot
\mathcal{F}\{\tilde{v}\}
\right)(t)
\end{equation}

Here, $\mathcal{F}$ and $\mathcal{F}^{-1}$ denote the Fourier and inverse Fourier transforms, and $\overline{(\cdot)}$ denotes complex conjugation. $\tilde{s}$ and $\tilde{v}$ are channel-aligned features obtained via lightweight projection.
This operation emphasizes shared periodic components and suppresses non-periodic noise, improving cross-modal consistency of the physiological signal.

\noindent\textbf{Confidence Gating.}
Due to varying waveform quality, the temporal representations from the video and STMap branches may differ, and direct information exchange can introduce interference. To address this, we introduce a confidence gating mechanism to adaptively control cross-modal information injection.

This mechanism assigns confidence scores at each time step $t$ to measure the reliability of modal features, including cross-modal consistency confidences $conf_{s\to v}(t)$ and $conf_{v\to s}(t)$, and waveform intensity confidences $conf_v(t)$ and $conf_s(t)$ for the video and STMap branches, respectively.

The cross-modal consistency confidence is computed from the cross-correlation sequence. Taking STMap→video as an example, let $r_{s\to v}(t)$ denote the correlation sequence. For each channel $c$, the confidence is defined as the ratio between its peak amplitude and its average amplitude over time:

\begin{equation}
\text{score}_{s \to v}(t)
=
\mathrm{median}_c
\left(
\frac{
\max_{\tau} \left| r_{s \to v, c}(\tau) \right|
}{
\epsilon
+
\frac{1}{T}
\sum_{\tau}
\left| r_{s \to v, c}(\tau) \right|
}
\right).
\label{eq:crossmodal_score_s2v}
\end{equation}

The resulting score is passed through  a sigmoid function to obtain the confidence value $conf_{s \to v}(t) \in [0,1]$, where higher values indicate stronger cross-modal consistency.

For modality-specific confidence, taking the video feature $v_{\text{global}}$ as an example, we first compute its Fourier transform to obtain the spectrum $V(f)$ and corresponding power spectrum $P(f)=|V(f)|^2$. The confidence score is then defined as the ratio of the maximum power to the average power, followed by channel-wise median aggregation.
\begin{equation}
\text{score}_v(t)
=
\mathrm{median}_c
\left(
\frac{
\max_f P_c(f)
}{
\epsilon + \mathrm{mean}_f \, P_c(f)
}
\right)
\label{eq:video_conf_score}
\end{equation}

This score is passed through a sigmoid function to obtain $conf_v(t)$, where clearer periodic peaks correspond to higher confidence. To improve temporal stability, all confidence scores are computed using a sliding window strategy and mapped back to form a smooth time-varying gating signal. These signals are used to weight the cross-modal features, enabling adaptive modulation of information injection and improving pulse waveform recovery.

For each time step $t$, the temporal gain coefficient for both branches are computed from the confidence scores as

\begin{align}
gain_v(t) &= conf_{s \to v}(t)\cdot conf_s(t), \quad
gain_s(t) = conf_{v \to s}(t)\cdot conf_v(t).
\end{align}
For each branch, the cross-modal matching features $r(c,t)$ are modulated by the corresponding temporal gains to produce residual feature sequences:
\begin{align}
v_{\text{res}}(c,t)
&=
\sigma\!\bigl(\alpha_{s2v}(c)\bigr)
\cdot
\bigl[
r_{s \to v}(c,t)\cdot gain_v(t)
\bigr],
\label{eq:video_residual} \\[2pt]
s_{\text{res}}(c,t)
&=
\sigma\!\bigl(\alpha_{v2s}(c)\bigr)
\cdot
\bigl[
r_{v \to s}(c,t)\cdot gain_s(t)
\bigr],
\label{eq:stmap_residual}
\end{align}
where $\alpha_{s2v}(c)$ and $\alpha_{v2s}(c)$ are learnable per-channel scaling weights.

Each residual signal is spatially broadcast using the corresponding attention map to form the injected feature increment:

\begin{equation}
V_{\text{inc}}(c,t,i,j)
=
A_{b,t,h,w}^{(S \to V)}\cdot v_{\text{res}}(c,t),
\quad
S_{\text{inc}}(c,h',t)
=
A_{b,t,h'}^{(V \to S)}\cdot s_{\text{res}}(c,t).
\label{eq:injection}
\end{equation}

The injected features are added back to the original features, yielding the updated outputs:
\begin{equation}
v_{\text{out}}=v+V_{\text{inc}},\quad
s_{\text{out}}=s+S_{\text{inc}}.
\end{equation}

Through the DCEB, the video and STMap branches exchange information to enhance feature fusion. The confidence-gated injection improves fusion robustness and provides more reliable features for final rPPG signal recovery.

\subsection{Structured Attention Fusion Decoder}
\label{sec:3.4Structured Attention Fusion Decoder}

To adaptively integrate the information from the video and STMap branches for reliable rPPG signal recovery, we introduce the structured attention fusion decoder to combine high-level semantic features from both branches.
The decoder takes the highest-level features $v_{T4}$ and $s_{T4}$ from the video and STMap branches, respectively. These features are projected into a shared embedding space to obtain the video token $\mathbf{v}_\text{tok} \in \mathbb{R}^{B \times T \times d}$ and the STMap token $\mathbf{s}_\text{tok} \in \mathbb{R}^{B \times T \times d}$. A state token and $N_q$ query tokens are introduced for structured attention fusion. The state token is initialized from the confidence values of both branches, providing global reliability context to the decoder.

Next, a multi-head self-attention layer with a structured attention mask $\mathbf{M}$ is applied for selective fusion. The mask controls information flow among tokens. STMap and video tokens attend to each other under the guidance of the state token for confidence-adaptive fusion. Query tokens attend only to STMap, video, and state tokens, but not to each other, while STMap and video tokens do not attend to query tokens. The state token attends to all tokens to regulate the global attention pattern.

\begin{equation}
\mathbf{M}_{i,j} =
\begin{cases}
-\lambda \cdot \max(0, \tau - \text{conf}_v), 
& \text{if } x_i \in \text{STMap},\ x_j \in \text{Video}, \\[2pt]
-\lambda \cdot \max(0, \tau - \text{conf}_s), 
& \text{if } x_i \in \text{Video},\ x_j \in \text{STMap}, \\[2pt]
-\infty, 
& \text{if } x_i, x_j \in \text{Query},\ i \ne j, \\[2pt]
-\infty,
& \text{if } x_j \in \text{Query},\ x_i \in \text{STMap or Video}, \\[2pt]
0, 
& \text{otherwise}.
\end{cases}
\label{eq:attention_mask_full}
\end{equation}

Here, $x_i$ and $x_j$ denote token types at positions $i$ and $j$, and $\lambda$ is a learnable scaling factor for soft gating. The attention computation follows the standard masked scaled dot-product attention:
\begin{equation}
\text{Attention}(Q, K, V) = \text{Softmax}\left(\frac{QK^\top}{\sqrt{d}} + \mathbf{M} \right) V.
\label{eq:masked_attention}
\end{equation}

This structured attention mechanism ensures that query tokens can focus on extracting the global temporal pattern across modalities; the state token controls the fusion weights between the two branches, acting as a gating mechanism; and video and STMap tokens achieve information complementarity without mutual interference.
The updated tokens then go through residual connections and feed-forward refinement:
\begin{equation}
X'=\text{LayerNorm}(X+\text{MHSA}(X)),\quad
X''=\text{LayerNorm}(X'+\text{FFN}(X'))
\end{equation}

Finally, the updated video token, STMap token, and query token $r_{\text{query}}(t)$ are aggregated to form the final fused representation:
\begin{equation}
\mathbf{f}_{\text{out}}(t) = \delta \cdot \mathbf{v}_\text{tok}(t) + (1-\delta)\cdot \mathbf{s}_\text{tok}(t) + \alpha \cdot \mathbf{r}_{\text{query}}(t),
\label{eq:fused_out}
\end{equation}
where $\delta$ and $\alpha$ are generated from the state token through a lightweight mapping network to control the fusion weights of each branch and the query output.

\section{Experiments}

\textbf{Datasets and Experimental Settings.} Detailed descriptions of the datasets, evaluation metrics, and implementation details are provided in Appendix A.

\subsection{Intra-Dataset Evaluation}

\begin{table*}[t]
  \caption{Intra-dataset testing results on  UBFC-RPPG~\cite{UBFCbobbia2019unsupervised}, PURE~\cite{PUREstricker2014non}, BUAA-MIHR~\cite{BUAAxi2020image}, and MMPD~\cite{tang2023mmpd}. The symbols $^{\triangle}$ and $^{\star}$ denote the traditional and deep learning-based methods, respectively. Best results are marked in \textbf{bold}.}
  \label{tab:main_results}
  \resizebox{\textwidth}{!}{
  \centering
  \small
  \begin{tabular}{@{}lcccccccccccc@{}}
    \toprule
    \multicolumn{1}{c}{\textbf{Method}} &
    \multicolumn{3}{c}{\textbf{UBFC-RPPG}} &
    \multicolumn{3}{c}{\textbf{PURE}} &
    \multicolumn{3}{c}{\textbf{BUAA-MIHR}} &
    \multicolumn{3}{c}{\textbf{MMPD}} \\
    \cmidrule(lr){2-4}\cmidrule(lr){5-7}\cmidrule(lr){8-10}\cmidrule(lr){11-13}
    & $\mathrm{MAE}\!\downarrow$ & $\mathrm{RMSE}\!\downarrow$ & $R\!\uparrow$
    & $\mathrm{MAE}\!\downarrow$ & $\mathrm{RMSE}\!\downarrow$ & $R\!\uparrow$
    & $\mathrm{MAE}\!\downarrow$ & $\mathrm{RMSE}\!\downarrow$ & $R\!\uparrow$
    & $\mathrm{MAE}\!\downarrow$ & $\mathrm{RMSE}\!\downarrow$ & $R\!\uparrow$\\
\midrule
    GREEN$^{\triangle}$~\cite{verkruysse2008remoteorigin}          & 19.73 & 31.00 & 0.37 & 10.09 & 23.85 & 0.34 &  6.89 & 10.39 & 0.60 & 21.68 & 27.69 & -0.01 \\
    ICA$^{\triangle}$\cite{poh2010advancements}            & 16.00 & 25.65 & 0.44 &  4.77 & 16.07 & 0.72 &  --   &  --   &  --   & 18.60 & 24.30 & 0.01 \\
    CHROM$^{\triangle}$\cite{de2013robust}          &  4.06 &  8.83 & 0.89 &  5.77 & 14.93 & 0.81 &  --   &  --   &  --   & 13.66 & 18.76 & 0.08 \\
    POS$^{\triangle}$\cite{poswang2016algorithmic}            &  4.08 &  7.72 & 0.92 &  3.67 & 11.82 & 0.88 &  --   &  --   &  --   & 12.36 & 17.71 & 0.18 \\
    Meta-rPPG$^{\star}$~\cite{lee2020meta}      &  5.97 &  7.42 & 0.57 &  2.52 &  4.63 & 0.98 &  --   &  --   &  --   &  --   &  --   &  --   \\
    PhysNet$^{\star}$~\cite{yu2019remotephysnet}        &  1.09 &  1.80 & 0.99 &  2.51 &  4.90 & 0.92 &  9.29 & 11.65 & 0.08 &  8.61 & 11.90 & 0.60 \\
    Efficientphys$^{\star}$~\cite{liu2023efficientphys}   &  1.41 &  1.81 & 0.99 &  4.75 &  9.39 & 0.99 & 16.09 & 16.80 & 0.14 & 13.47 & 21.32 & 0.21 \\
    PhysFormer$^{\star}$~\cite{yu2022physformer}     &  0.92 &  2.46 & 0.99 &  1.10 &  1.75 & 0.99 &  --   &  --   &  --   & 11.99 & 18.41 & 0.18 \\
    Contrast-Phys+$^{\star}$~\cite{sun2024Contrast-phys+} &  0.21 &  0.80 & 0.99 &  0.48 &  0.95 & 0.99 &  --   &  --   &  --   &  --   &  --   &  --   \\
    PhysDiff$^{\star}$~\cite{qian2025physdiff}       &  0.33 &  0.57 & 0.99 &  0.29 &  0.54 & 0.99 &  --   &  --   &  --   &  7.17 &  9.63 & 0.78 \\
    RhythmFormer$^{\star}$~\cite{zou2025rhythmformer}   &  0.50 &  0.78 & 0.99 &  0.27 &  \textbf{0.47} & 0.99 &  9.19 & 11.93 & -0.10 &  \textbf{4.69} & 11.31 & 0.60 \\
    LST-rPPG$^{\star}$~\cite{li2025lst}       &  0.16 &  0.57 & 0.99 &  0.32 &  0.62 & 0.99 &  --   &  --   &  --   &  --   &  --   &  --   \\
    \textbf{PhysNeXt (Ours)$^{\star}$} 
                   & \textbf{0.15} & \textbf{0.33} & \textbf{0.99}
                   & \textbf{0.24} & 0.54 & \textbf{0.99}
                   & \textbf{5.80} & \textbf{6.94} & \textbf{0.62}
                   & 5.06 & \textbf{6.65} & \textbf{0.87} \\
    \bottomrule
  \end{tabular}
  }
  \vspace{-2em}
\end{table*}

\cref{tab:main_results} reports the intra-dataset evaluation results of the proposed PhysNeXt on four standard benchmarks, namely UBFC-RPPG~\cite{UBFCbobbia2019unsupervised}, PURE~\cite{PUREstricker2014non}, BUAA-MIHR~\cite{BUAAxi2020image}, and MMPD~\cite{tang2023mmpd}, together with systematic comparisons against representative state-of-the-art methods~\cite{liu2023efficientphys,zou2025rhythmformer,li2025lst}. The results show that PhysNeXt achieves the best or near-best performance across almost all metrics, demonstrating its strong robustness.

On the UBFC-RPPG dataset, PhysNeXt achieves the lowest MAE (0.15 ) and RMSE (0.33), as well as the highest Pearson correlation coefficient (R = 0.99), significantly outperforming advanced methods such as RhythmFormer~\cite{zou2025rhythmformer}, PhysFormer~\cite{yu2022physformer}, and PhysDiff~\cite{qian2025physdiff}. This indicates that our model can accurately fit heart-rate waveforms in relatively stable desktop recording scenarios.
On the PURE dataset, PhysNeXt continues to deliver leading performance with an MAE of 0.24, RMSE of 0.54, and R of 0.99, outperforming RhythmFormer~\cite{zou2025rhythmformer} (MAE 0.27) and PhysDiff~\cite{qian2025physdiff} (MAE 0.29). This strong performance suggests the advantage of the proposed model design in handling temporal disturbances and maintaining stable pulse rhythm modeling.
On the low-illumination BUAA-MIHR dataset, PhysNeXt achieves RMSE = 5.80 and R = 0.62, which are significantly better than those of competing methods. These results indicate that PhysNeXt maintains robust performance under low-light conditions and can effectively recover physiological signals in challenging environments.
On the challenging MMPD dataset, PhysNeXt achieves an RMSE of 6.65 and R of 0.87, outperforming PhysDiff~\cite{qian2025physdiff} and RhythmFormer~\cite{zou2025rhythmformer}. This result highlights the strong robustness and effectiveness of PhysNeXt in complex real-world scenarios.

\subsection{Cross-Dataset Evaluation}
\textbf{Multi-Domain Generalization.} 
\cref{tab:cross_domain} presents the cross-dataset generalization performance under a leave-one-out protocol. PhysNeXt achieves strong results across all target domains. When UBFC-RPPG\cite{UBFCbobbia2019unsupervised} is used as the target, it obtains MAE = 3.81, RMSE = 7.63, and R = 0.92, outperforming existing methods. On PURE\cite{PUREstricker2014non}, it achieves R = 0.82 with reduced MAE and RMSE, demonstrating robustness to motion and illumination changes. PhysNeXt also maintains competitive performance on the low-light BUAA-MIHR dataset\cite{BUAAxi2020image} (R = 0.94, RMSE = 3.88) and the challenging MMPD dataset\cite{tang2023mmpd} (MAE = 10.89, RMSE = 17.37, R = 0.30), indicating strong generalization ability.

These cross-dataset results validate the strong generalization capability of PhysNeXt across different devices, lighting conditions, motion disturbances, and subject populations. Benefiting from its dual-stream architecture that jointly models video and STMap representations, together with waveform matching and confidence-gated interaction, PhysNeXt can adaptively select the most reliable temporal signal sources while suppressing redundant and noisy information. 

\begin{table*}[t]
\caption{Multi-domain generalization evaluation under the leave-one-out protocol. 
U=UBFC-RPPG~\cite{UBFCbobbia2019unsupervised}, P=PURE~\cite{PUREstricker2014non}, B=BUAA-MIHR~\cite{BUAAxi2020image}, and M=MMPD~\cite{tang2023mmpd}.
Best results are marked in \textbf{bold}.}
\label{tab:cross_domain}
\resizebox{\textwidth}{!}{
\centering
\small

\begin{tabular}{@{}lcccccccccccc@{}}

\toprule
\multicolumn{1}{c}{\textbf{Method}} &
\multicolumn{3}{c}{\textbf{Other$\rightarrow$U}} &
\multicolumn{3}{c}{\textbf{Other$\rightarrow$P}} &
\multicolumn{3}{c}{\textbf{Other$\rightarrow$B}} &
\multicolumn{3}{c}{\textbf{Other$\rightarrow$M}} \\
\cmidrule(lr){2-4}\cmidrule(lr){5-7}\cmidrule(lr){8-10}\cmidrule(lr){11-13}
& $\mathrm{MAE}\!\downarrow$ & $\mathrm{RMSE}\!\downarrow$ & $R\!\uparrow$
& $\mathrm{MAE}\!\downarrow$ & $\mathrm{RMSE}\!\downarrow$ & $R\!\uparrow$
& $\mathrm{MAE}\!\downarrow$ & $\mathrm{RMSE}\!\downarrow$ & $R\!\uparrow$
& $\mathrm{MAE}\!\downarrow$ & $\mathrm{RMSE}\!\downarrow$ & $R\!\uparrow$ \\
\midrule
    NEST~\cite{NESTlu2023neuron}           & 17.24 & 20.58 & 0.06 & 19.26 & 26.11 & -0.06 &  9.19 & 12.38 & 0.19 & 13.97 & 18.20 & 0.15 \\
    GREIP~\cite{greipzhang2025advancing}          & 17.50 & 20.42 & 0.17 &  5.07 & 14.50 & 0.78 &  7.94 & 10.93 & -0.03 & 13.02 & 17.11 & 0.14 \\
    EfficientPhys~\cite{liu2023efficientphys}  & 12.87 & 18.80 & 0.19 &  7.15 & 15.04 & 0.23 & 32.30 & 34.00 & -0.03 & 12.87 & 18.80 & 0.19 \\
    PhysFormer~\cite{yu2022physformer}      & 10.29 & 18.13 & 0.60 & 19.75 & 24.30 & 0.24 & 22.09 & 26.21 & 0.03 & 13.90 & 19.30 & 0.06 \\
    PhysNet~\cite{yu2019remotephysnet}        & 13.83 & 23.66 & 0.35 & 33.23 & 35.25 & -0.15 & 12.75 & 16.37 & 0.08 & 13.37 & \textbf{16.64} & 0.29 \\
    RhythmFormer~\cite{zou2025rhythmformer}   & 14.71 & 22.49 & 0.43 & 25.76 & 25.76 & 0.04 &  6.04 & 10.84 & 0.42 & 16.14 & 20.50 & -0.11 \\
    \textbf{PhysNeXt(Ours)} 
                  & \textbf{3.81} & \textbf{7.63} & \textbf{0.92}
                  & \textbf{5.05} & 13.08 & \textbf{0.82}
                  & \textbf{2.94} & \textbf{3.88} & \textbf{0.94}
                  & \textbf{10.89} & 17.37 & \textbf{0.30} \\
    \bottomrule
  \end{tabular}
  }
  \vspace{-1em}

\end{table*}
\begin{table*}[t]
\caption{Limited-source domain generalization evaluation on BUAA-MIHR. 
Models are trained on limited source domains.
U=UBFC-RPPG~\cite{UBFCbobbia2019unsupervised}, P=PURE~\cite{PUREstricker2014non}, B=BUAA-MIHR~\cite{BUAAxi2020image}, and M=MMPD~\cite{tang2023mmpd}. Best results are marked in \textbf{bold}.}
\label{tab:limited_source_dg}
\resizebox{\textwidth}{!}{
\centering
\small
\vspace{-1em}
\begin{tabular}{@{}lcccccccccccc@{}}

\toprule
\multicolumn{1}{c}{\textbf{Method}} &
\multicolumn{3}{c}{\textbf{U+P$\rightarrow$B}} &
\multicolumn{3}{c}{\textbf{M+P$\rightarrow$B}} &
\multicolumn{3}{c}{\textbf{U+M$\rightarrow$B}} &
\multicolumn{3}{c}{\textbf{Average}} \\
    \cmidrule(lr){2-4}\cmidrule(lr){5-7}\cmidrule(lr){8-10}\cmidrule(lr){11-13}
    & $\mathrm{MAE}\!\downarrow$ & $\mathrm{RMSE}\!\downarrow$ & $R\!\uparrow$
    & $\mathrm{MAE}\!\downarrow$ & $\mathrm{RMSE}\!\downarrow$ & $R\!\uparrow$ 
    & $\mathrm{MAE}\!\downarrow$ & $\mathrm{RMSE}\!\downarrow$ & $R\!\uparrow$ 
    & $\mathrm{MAE}\!\downarrow$ & $\mathrm{RMSE}\!\downarrow$ & $R\!\uparrow$ \\
    \midrule
        PhysNet~\cite{yu2019remotephysnet}        & 15.34 & 21.48 & -0.29 & 20.97 & 24.75 & 0.01 & 11.40 & 16.72 & 0.14 & 15.90 & 20.98 & -0.05 \\
    PhysFormer~\cite{yu2022physformer}      & 18.23 & 22.17 & 0.07 & 14.86 & 18.26 & 0.03 & 10.87 & 16.20 & 0.08 & 14.65 & 18.88 & 0.06 \\
    EfficientPhys~\cite{liu2023efficientphys}  &  4.60 &  8.06 & 0.72 &  4.15 &  7.14 & 0.77 &  3.00 &  5.18 & 0.89 &  3.92 &  6.79 & 0.79 \\
    RhythmFormer~\cite{zou2025rhythmformer}   &  3.90 &  6.51 & 0.82 &  4.32 &  6.70 & 0.82 &  6.20 & 11.23 & 0.49 &  4.81 &  8.15 & 0.71 \\
    NEST~\cite{NESTlu2023neuron}           & 11.19 & 14.36 & -0.12 &  8.67 & 11.38 & 0.05 &  9.14 & 12.09 & -0.04 &  9.67 & 12.60 & -0.04 \\
    GREIP~\cite{greipzhang2025advancing}          & 10.25 & 13.08 & 0.02 &  8.69 & 11.34 & 0.11 & 11.40 & 14.27 & 0.04 & 10.11 & 12.90 & 0.06 \\
    \textbf{PhysNeXt (Ours)} 
                   & \textbf{2.67} & \textbf{3.50} & \textbf{0.97}
                   & \textbf{2.88} & \textbf{3.91} & \textbf{0.94}
                   & \textbf{2.83} & \textbf{3.73} & \textbf{0.95}
                   & \textbf{2.79} & \textbf{3.71} & \textbf{0.95} \\
    \bottomrule
  \end{tabular}
  }
  \vspace{-2em}
\end{table*}

\begin{table*}[t]
\caption{Limited-source domain generalization evaluation on MMPD. 
Models are trained on limited combinations of source domains. Best results are marked in \textbf{bold}.}
\label{tab:limited_source_dg_mmpd}
\resizebox{\textwidth}{!}{
\centering
\small
\setlength{\tabcolsep}{1.0pt}
\renewcommand{\arraystretch}{1.05}
\begin{tabular}{@{}lcccccccccccc@{}}

\toprule
\multicolumn{1}{c}{\textbf{Method}} &
\multicolumn{3}{c}{\textbf{U+P$\rightarrow$M}} &
\multicolumn{3}{c}{\textbf{B+P$\rightarrow$M}} &
\multicolumn{3}{c}{\textbf{U+B$\rightarrow$M}} &
\multicolumn{3}{c}{\textbf{Average}} \\
\cmidrule(lr){2-4}\cmidrule(lr){5-7}\cmidrule(lr){8-10}\cmidrule(lr){11-13}
& $\mathrm{MAE}\!\downarrow$ & $\mathrm{RMSE}\!\downarrow$ & $R\!\uparrow$
& $\mathrm{MAE}\!\downarrow$ & $\mathrm{RMSE}\!\downarrow$ & $R\!\uparrow$
& $\mathrm{MAE}\!\downarrow$ & $\mathrm{RMSE}\!\downarrow$ & $R\!\uparrow$
& $\mathrm{MAE}\!\downarrow$ & $\mathrm{RMSE}\!\downarrow$ & $R\!\uparrow$\\
\midrule
PhysNet~\cite{yu2019remotephysnet}        & 11.00 & 17.30 & 0.28  & 13.20 & 16.70 & \textbf{0.23} & 13.50 & 17.00 & 0.09 & 12.57 & 17.00 & 0.20 \\
PhysFormer~\cite{yu2022physformer}      & 11.40 & 17.50 & 0.23  & 13.90 & 18.60 & 0.21 & 13.20 & \textbf{16.50} & 0.12 & 12.83 & 17.53 & 0.19 \\
EfficientPhys~\cite{liu2023efficientphys}  & 11.80 & 18.90 & 0.22  & \textbf{11.90} & 18.50 & 0.21 & 15.50 & 20.80 & 0.03 & 13.07 & 19.40 & 0.15 \\
RhythmFormer~\cite{zou2025rhythmformer}   & 10.50 & 16.72 & 0.28  & 13.98 & 19.46 & 0.12 & 12.57 & 17.45 & 0.15 & 12.35 & 17.88 & 0.18 \\
NEST~\cite{NESTlu2023neuron}           & 14.01 & 18.12 & 0.12  & 12.82 & 17.26 & 0.22 & 13.29 & 16.58 & 0.16 & 13.37 & 17.32 & 0.17 \\
GREIP~\cite{greipzhang2025advancing}          & 12.90 & 16.54 & 0.11 & 12.81 & 17.09 & 0.12 & 14.96 & 18.13 & -0.05 & 13.56 & 17.25 & 0.06 \\
\textbf{PhysNeXt (Ours)}
              & \textbf{9.79} & \textbf{16.46} & \textbf{0.37}
              & 12.49 &\textbf{15.49} & 0.14
              & \textbf{10.61} & 16.70 & \textbf{0.32}
              & \textbf{10.96} &\textbf{16.22}  & \textbf{0.28} \\
\bottomrule
\end{tabular}
}
\vspace{-2em}
\end{table*}

\noindent\textbf{Limited-Source Domain Generalization.} Under limited source-domain conditions, we further evaluate the cross-dataset generalization capability of PhysNeXt. 
As shown in \cref{tab:limited_source_dg}, when BUAA-MIHR~\cite{BUAAxi2020image} is used as the target domain under three configurations (U+P → B, M+P → B, and U+M → B), PhysNeXt achieves MAE values of 2.67, 2.88, and 2.83, RMSE values of 3.50, 3.91, and 3.73, and Pearson correlation coefficients (R) of 0.97, 0.94, and 0.95, respectively. PhysNeXt consistently achieves superior performance across all evaluation metrics. 
Similarly, as reported in \cref{tab:limited_source_dg_mmpd}, when MMPD~\cite{tang2023mmpd} serves as the target domain, PhysNeXt achieves MAE values of 9.79, 12.49, and 10.61, RMSE values of 16.46, 15.49, and 16.70, and R values of 0.37, 0.14, and 0.32 under source-domain combinations (U+P → M, B+P → M, and U+B → M), respectively. PhysNeXt attains the best performance in most configurations, and its average results are clearly superior to those of other deep learning models. Given the complex motion patterns, illumination changes, and skin tone variations present in the MMPD dataset, these results demonstrate that PhysNeXt maintains strong cross-domain performance even under multi-factor interference.

Overall, these results demonstrate that PhysNeXt exhibits strong out-of-domain generalization capability even when training data are limited and source-domain differences are significant.
This advantage comes from the proposed dual-stream modeling structure for collaborative multi-modal encoding, as well as the structured attention mechanism introduced in the fusion decoder. These designs improve the modeling of key pulse signals and reduce reliance on dataset-specific appearance features, enabling stable and efficient cross-domain prediction.

\subsection{Ablation Study}

\textbf{Study of Different Architectures and Modules.}
To evaluate the effectiveness of the overall architecture and key modules, we conduct ablation studies on the UBFC~\cite{UBFCbobbia2019unsupervised} and MMPD~\cite{tang2023mmpd} datasets. Using only a single branch (STMap or video) leads to degraded performance, indicating that both representations provide complementary information. The STMap-only model shows much higher error on UBFC, while the video-only model suffers from reduced robustness on the noisy MMPD dataset.
We further examine the cross-modal interaction mechanism. Restricting information exchange to a single direction (STMap → Video or Video → STMap) results in performance drops, demonstrating the importance of bidirectional interaction. In particular, the Video → STMap setting achieves RMSE = 11.98 on MMPD, compared with 6.65 for the full model.
We also evaluate the contribution of individual components. Removing the DCEB increases RMSE to 10.97 on MMPD. Removing the structured attention decoder or the spatio-temporal difference modeling unit also leads to clear performance degradation. 
Overall, these results show that each component contributes to performance improvement, confirming the effectiveness of the PhysNeXt design.
\begin{figure}[t]
  \centering
  \includegraphics[width=\linewidth]{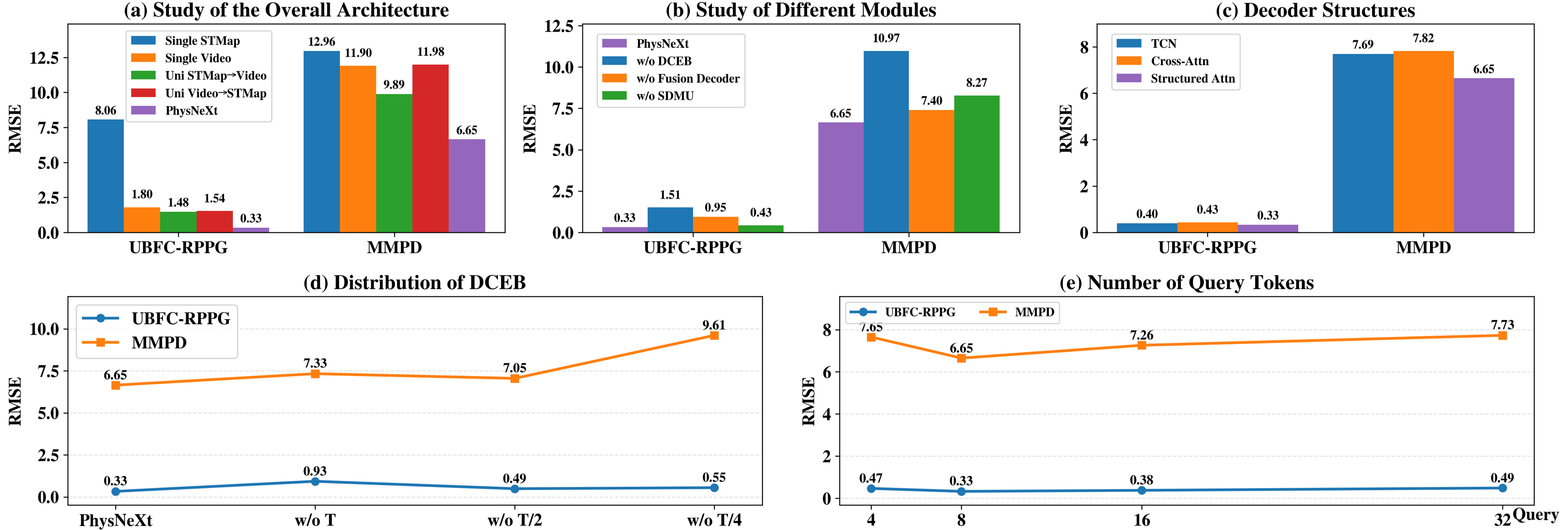}
  \caption{Ablations on PhysNeXt architecture, covering dual-stream configurations, temporal exchange module (DCEB), decoder structure, and the number of query tokens.}
  \label{fig:ablation_results}
  \vspace{-2em}
\end{figure}

\noindent\textbf{Decoder Structures.}
We compare the impact of different decoder structures on model performance, including temporal convolutional network (TCN), cross-attention, and our proposed structured attention mechanism. Experimental results show that structured attention achieves the best RMSE score of 0.33 on the UBFC dataset~\cite{UBFCbobbia2019unsupervised}, outperforming TCN (0.40) and cross-attention (0.43). On the more challenging MMPD dataset~\cite{tang2023mmpd}, structured attention also achieves superior performance with an RMSE of 6.65. 
These results suggest that, compared to TCN and standard attention mechanisms, structured attention is better suited for handling cross-modal fusion features and focusing on key temporal patterns, thereby improving prediction accuracy and robustness.

\vspace {0.2em}
\noindent\textbf{Distribution of DCEB.}
To investigate the impact of the Dual-Stream Confidence-Gated Exchange Block (DCEB), we conduct ablation experiments on its distribution across different temporal scales (T, T/2, and T/4) on the UBFC~\cite{UBFCbobbia2019unsupervised} and MMPD~\cite{tang2023mmpd} datasets. Each time, one exchange block is removed while keeping the others unchanged.
As shown in \cref{fig:ablation_results}, the full model with exchange blocks at all three scales achieves the best performance, indicating the importance of multi-scale interaction. Removing the shallow exchange block leads to clear degradation, especially on UBFC. Removing the middle block causes a smaller drop, while removing the deep block significantly affects performance on MMPD, showing its importance in complex scenarios.
Overall, these results demonstrate that exchange blocks at different temporal scales contribute jointly to effective cross-modal fusion and robust physiological signal estimation.

\vspace {0.2em}
\noindent\textbf{Number of Query Tokens.} To evaluate the impact of the number of query tokens in the structured attention decoder, we compare four settings: 4, 8, 16, and 32. Experiments reveal that performance is optimal at 8 tokens, yielding the lowest RMSE of 0.33 on UBFC~\cite{UBFCbobbia2019unsupervised} and 6.65 on MMPD~\cite{tang2023mmpd}. This indicates that while decreasing the token count from 32 to 8 helps the model focus on key information and suppress redundant interference, further reducing it to 4 limits representational capacity. Consequently, considering both performance and stability, we ultimately chose 8 query tokens as the default configuration.

\subsection{Visualization and Analysis}
\begin{figure}[t]
  \centering
  \includegraphics[width=\linewidth]{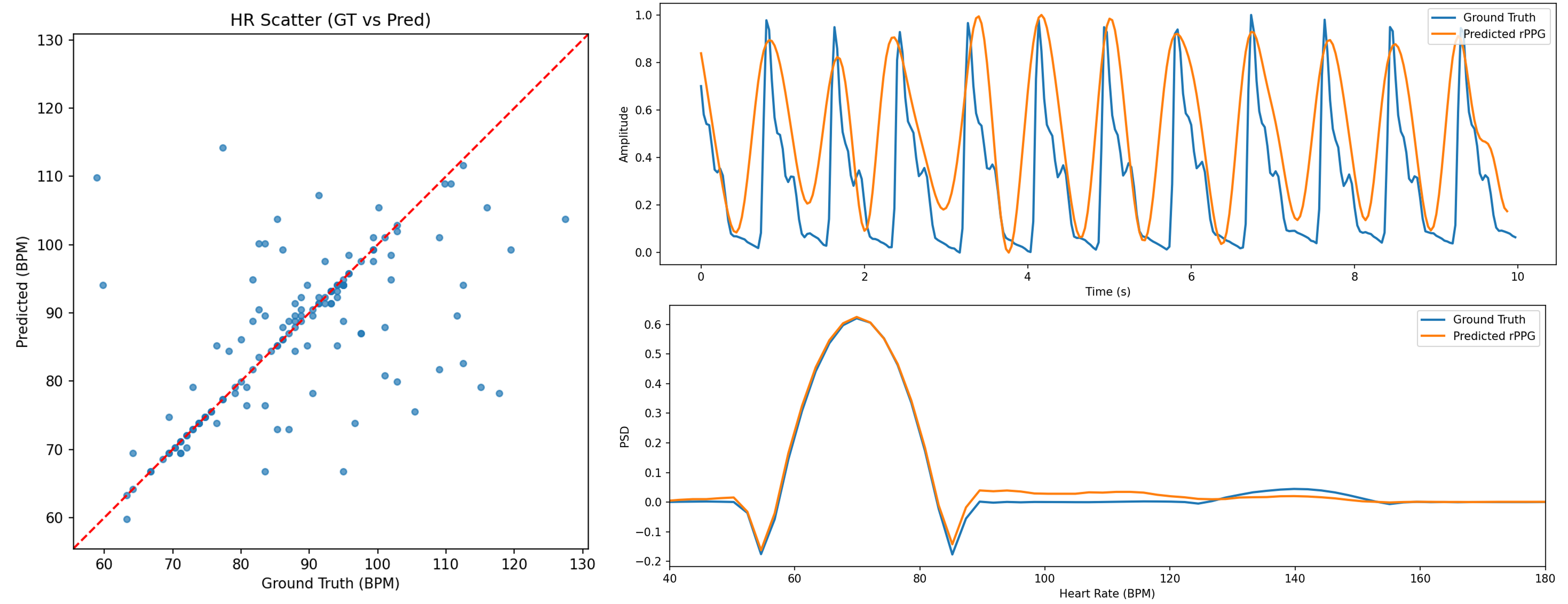}
  \caption{Visualization results on the MMPD dataset. Left: Scatter plot of predicted heart rate versus ground truth. Upper right: Time-domain comparison between the predicted and ground truth rPPG waveforms. Lower right: Frequency-domain power spectral density (PSD) comparison.}
  \label{fig:visual}
\vspace{-2em}
\end{figure}
\textbf{Visualization.} 
\cref{fig:visual} shows the heart rate estimation results of PhysNeXt on the MMPD dataset. The scatter plot compares the predicted and ground truth heart rates, with the red dashed line representing the ideal prediction. Most points are near this line, indicating good accuracy. The time-domain plot compares the predicted rPPG signal with the ground truth, showing high consistency in rhythm patterns, demonstrating the model's pulse waveform recovery capability. The PSD plot shows the energy distribution across frequencies, with closely aligned peak locations in both predicted and true signals, further validating the model’s accurate representation of key physiological rhythms. Overall, this visualization confirms the robustness and accuracy of PhysNeXt in rPPG signal recovery and heart rate estimation.

\begin{wraptable}{r}{0.4\textwidth}
\vspace{-25pt} 
\caption{Efficiency of different rPPG methods.}
\label{tab:efficiency}
\centering 
\resizebox{0.4\textwidth}{!}{
\begin{tabular}{lcc}
\toprule
\textbf{Method} & \textbf{Params (M)} & \textbf{MACs (G)} \\
\midrule
DeepPhys\cite{chen2018deepphys}               & 7.50 & 96.0  \\
EfficientPhys~\cite{liu2023efficientphys}     & 7.40 & 45.6  \\
PhysFormer\cite{yu2022physformer}             & 7.38 & 40.5  \\
RhythmFormer\cite{zou2025rhythmformer}        & 4.21 & 28.8  \\
Contrast-Phys+\cite{sun2024Contrast-phys+}    & 0.85 & 145.7 \\
PhysNet\cite{yu2019remotephysnet}             & 0.77 & 56.1  \\
PhysNeXt (Ours)                               & 4.66 & 58.1  \\
\bottomrule
\end{tabular}%
}
\vspace{-25pt} %
\end{wraptable}

\vspace {0.2em}
\noindent\textbf{Efficiency Analysis.}
The efficiency analysis shows that PhysNeXt achieves a balance between model size (4.66M parameters) and computational overhead (58.1G MACs). Compared to methods like DeepPhys~\cite{chen2018deepphys}, EfficientPhys~\cite{liu2023efficientphys}, and PhysFormer~\cite{yu2022physformer}, PhysNeXt reduces the model size while maintaining similar or better efficiency. Additionally, compared to Contrast-Phys+~\cite{sun2024Contrast-phys+} (0.85M / 145.7G MACs), PhysNeXt offers a more efficient design, balancing compactness and inference speed, and avoiding bottlenecks from redundant computations.

\section{Conclusions}

In this paper, we propose PhysNeXt, a dual-stream rPPG estimation model that takes both video frames and STMap representations as inputs. 
Specifically, the proposed Spatio-Temporal Difference Modeling Unit enhances the STMap branch’s ability to model subtle skin color variations induced by heartbeats. The proposed Dual-Stream Confidence-Gated Exchange Block enables effective cross-modal information interaction between the video and STMap branches, ensuring complementary feature learning. The proposed Structured Attention Decoder further fuses the two streams and significantly improves the accuracy of rPPG signal recovery. 
Extensive experiments and ablation studies show that PhysNeXt significantly outperforms existing methods on multiple rPPG datasets.

\bibliographystyle{splncs04}
\bibliography{main}
\end{document}